\documentclass[letterpaper]{article} 
\usepackage[draft]{aaai25}  
\usepackage{times}  
\usepackage{helvet}  
\usepackage{courier}  
\usepackage[hyphens]{url}  
\usepackage{graphicx} 
\usepackage{amsmath}
\usepackage{amsthm}
\usepackage{booktabs}
\usepackage{multirow}
\usepackage{algorithm}
\usepackage{algorithmic}
\usepackage{subcaption}

\usepackage{tikz}
\usetikzlibrary{shapes.geometric, arrows}

\usepackage{natbib}  
\usepackage{caption} 
\usepackage{amssymb}
\usepackage{xcolor}
\usepackage{newfloat}
\usepackage{listings}
\DeclareCaptionStyle{ruled}{labelfont=normalfont,labelsep=colon,strut=off} 
\floatstyle{ruled}
\newfloat{listing}{tb}{lst}{}
\floatname{listing}{Listing}
\title{Learning Valid Dual Bounds in Constraint Programming: \\
 Boosted Lagrangian Decomposition with Self-Supervised Learning}
\author{
    Swann Bessa,\textsuperscript{\rm 1,2}
    Darius Dabert,\textsuperscript{\rm 1, 2}
    Max Bourgeat,\textsuperscript{\rm 1}\\
    Louis-Martin Rousseau,\textsuperscript{\rm 1}
    Quentin Cappart\textsuperscript{\rm 1}}

\affiliations {
    \textsuperscript{\rm 1} Polytechnique Montréal, Montreal, Canada\\
    \textsuperscript{\rm 2} Ecole Polytechnique, Palaiseau, France\\
    \{swann.bessa, darius.dabert\}@polytechnique.edu \\
    \{max.bourgeat,  louis-martin.rousseau, quentin.cappart\}@polymtl.ca
}

\newcommand{\qcap}[1]{{{\textcolor{teal}{\textbf{[QC:} {#1}\textbf{]}}}}}

\newcommand{\yes}[1]{{{\textcolor{teal}{\textbf{[YES]}}}}}

\begin{document}

\maketitle

\begin{abstract}
Lagrangian decomposition (LD) is a relaxation method that provides a 
dual bound for constrained optimization problems by decomposing them into more manageable sub-problems. 
This bound can be used in branch-and-bound algorithms to prune the search space effectively.
In brief, a vector of Lagrangian multipliers is associated with each sub-problem, and an iterative procedure (e.g., a sub-gradient optimization) adjusts these multipliers to find the tightest bound.
Initially applied to integer programming, 
Lagrangian decomposition also had success in constraint programming due to
its versatility and the fact that global constraints provide natural sub-problems. However, the non-linear and combinatorial nature of sub-problems in constraint programming makes it computationally intensive to optimize the Lagrangian multipliers with sub-gradient
methods at each node of the tree search. 
This currently limits the practicality of LD as a general bounding mechanism for constraint programming. To address this challenge,
we propose a self-supervised learning approach that leverages neural networks to generate multipliers directly, yielding tight bounds.
This approach significantly reduces the number of sub-gradient optimization steps required, 
enhancing the pruning efficiency and reducing the execution time of constraint programming solvers.
This contribution is one of the few that leverage learning to enhance
bonding mechanisms on the dual side,
a critical element in the design of combinatorial solvers.
To our knowledge, this work presents 
the first generic method for learning valid dual bounds in constraint programming.
We validate our approach on two challenging combinatorial problems: 
The multi-dimensional knapsack problem and the shift scheduling problem. The  results show that our 
approach can solve more instances than the standard application of LD to constraint programming, 
reduce execution time by more than half, and has promising generalization ability through fine-tuning.
\end{abstract}

\section{Introduction}

Combinatorial optimization has applications across diverse fields, such as aerospace, transportation planning, scheduling, and economics. The objective is to identify an optimal solution from a finite set of possibilities. A significant challenge in combinatorial optimization is the state-space explosion problem: the number of potential solutions grows exponentially with the size of the problem, rendering large-scale problem-solving intractable.
\textit{Constraint programming} (CP) is a versatile tool for addressing combinatorial optimization problems~\citep{rossi2006handbook} as it is not confined to linear models, unlike mixed-integer programming (MIP). It can theoretically address any combinatorial problem, including those with non-linear constraints or objectives.
Standard CP solvers rely on a search procedure that exhaustively yet efficiently enumerates all possible assignments of values to variables until the optimal solution is found. 

The search is commonly conducted depth-first, combined with \textit{branch-and-bound} techniques. The search backtracks to the previous decision point when an infeasible solution is encountered due to an empty domain. Through this procedure, and given that the entire search space is explored, the final solution found is proven to be optimal. Unfortunately, this tree search grows exponentially with the number of variables, making a complete enumeration of solutions still intractable for large problems. To address this challenge, CP search is first enhanced through a mechanism called \textit{propagation}, which reduces the number of possible combinations and, consequently, the size of the tree search. Given the current domains of the variables and a constraint, propagation eliminates values from the domains that cannot be part of the final solution due to constraint violations. This process is repeated with each domain change and for each constraint until no more values can be removed from the domains.

Given its reliance on branch-and-bound, a first design choice in CP is the \textit{branching} strategy, which dictates how the search space is explored. Well-designed heuristics are more likely to discover promising solutions, whereas poor heuristics may lead the search into unproductive regions of the solution space. The selection of an appropriate branching strategy is generally non-trivial, and its automatic design with learning methods has been extensively studied by the research community~\cite{michel2012activity,song2022learning,marty2023learning}.
A second critical choice, though much less explored through the lens of learning, is the \textit{bounding} strategy. Most existing CP solvers rely on a straightforward mechanism. When a feasible solution is found, a new constraint is added to ensure the following solution is better than the current one. This new constraint can enable more propagation. Unfortunately, the filtering provided by this approach is often relatively poor. Furthermore, this mechanism only allows the computation of a primal bound and does not offer dual bounds acting as a certificate of optimality.
A significant shortcoming of current CP technologies is the lack of a general and efficient dual-bounding mechanism for constraint programs, unlike MIP solvers, which rely on the linear relaxation. This limitation is one of the main reasons why CP solvers are often less competitive than MIP solvers for many combinatorial problems. This paper proposes to tackle this limitation.

\section{Related Work}

\subsubsection{Cost-Based Domain Filtering.} 
To our knowledge, \citet{focacci1} were the first to leverage bounding methods from mathematical programming to improve CP solvers generically.
They propose to embed in constraints an optimization component consisting of a 
linear program representing a linear relaxation of the constraint itself. 
This is commonly referred to as \textit{cost-based domain filtering} and directly enables the computation of a dual bound when solving the relaxed problem.
However, a drawback of cost-based domain filtering is that the bound is local to each constraint for which a custom relaxation module needs to be developed. At the time of writing, the global constraints catalog reports more than 400 different existing global constraints~\citep{beldiceanu2007global}, but only two of them have been considered in this work: \textsc{allDifferent} and \textsc{path}.
This naturally led researchers to implement this idea for other global constraints \citep{fahle2002cost, sellmann2003cost, houndji2017weighted}, or to leverage different types of relaxations. For instance, \citet{benchimol2012improved} proposed using the Held and Karp relaxation to improve the filtering of the \textsc{WeightedCircuit} constraint.

\subsubsection{CP-based Lagrangian Relaxation.}

Cost-based domain filtering has a second drawback: examining only one constraint at a time risks significantly underestimating the actual objective cost. This occurs because the mechanism often assumes it is possible to improve the objective, which can be impossible due to other constraints. This approach acts locally and lacks a global view of the problem.
One solution is to soften all constraints and incorporate them into the objective function with a penalty term, referred to as \textit{Lagrangian multipliers}. This approach simplifies the problem. In CP, this procedure was introduced by \citet{sellmann2001cp} and is commonly known as \textit{CP-based Lagrangian relaxation}.
\citet{sellmann2004theoretical} later proved that suboptimal Lagrangian multipliers can have stronger filtering abilities than optimal ones, which is advantageous as it allows for better filtering with less computational power. This idea has subsequently been used to solve various problems~\citep{sellmann2003constraint, cambazard2010hybrid, isoart2020adaptive, boudreault2021improved}, involving more and more constraints~\citep{menana2009sequencing, cambazard2015new, berthiaume2024local}, and has shown promise in improving the performance of CP solvers~\citep{fontaine2014constraint}.

    A drawback of CP-based Lagrangian relaxation is that the initial problem loses part of its original structure as the combinatorial constraints are removed from the constraint store and moved into the objective function. Additionally, the penalized constraints must be either linear or easily linearizable, which is not often true for most problems. 

\subsubsection{Lagrangian Decomposition in CP.} 

This limitation has progressively led researchers to consider \textit{Lagrangian decomposition}~\cite {guignard1987lagrangean} as an alternative bounding scheme, with the advantage of applying it to various constraints.  Lagrangian decomposition breaks down the initial problem into a master problem and several independent sub-problems, each of which is more manageable to solve. A vector of Lagrangian multipliers is associated with each sub-problem, and the master problem iteratively adjusts these multipliers to identify the optimal penalties, thereby guiding the solution process toward optimality. This is typically achieved through a sub-gradient procedure~\cite {shor2012minimization}, wherein each sub-problem must be solved exactly once per sub-gradient iteration.
\citet{cronholm2004strong} were the first to apply this concept within the context of constraint programming, proposing a specific algorithm for the multicast network design problem. A decade later, \citet{ha2015general} advanced this idea to develop an automatic bounding mechanism for constraint programming. Concurrently, \citet{bergman2015improved} suggested utilizing Lagrangian decomposition for constraints that can be represented by decision diagrams~\cite{andersen2007constraint}. Both studies underscored the potential of Lagrangian decomposition as an automatic dual-bounding mechanism for constraint programming.

However, two significant challenges arise with Lagrangian decomposition. First, a specialized and efficient algorithm is required to solve the sub-problems. Second, it incurs substantial computational overhead as each sub-problem must be solved at every sub-gradient iteration. While the former challenge has been addressed by \citet{chu2016lagrangian}, who proposed solving the sub-problems via a search procedure rather than a specialized propagator, the latter remains unresolved. To our knowledge, the efficient utilization of Lagrangian decomposition for CP has not been further investigated and remains an open research question for the practical application of this approach.

\subsubsection{Learning Valid Dual Bounds.}

In another context, machine learning has been considered to enhance the solving process of
branch-and-bound solvers, both for integer programming~\cite{khalil2016learning} and constraint programming~\cite{cappart2021combining}. We refer to the surveys of~\citet{bengio2021machine} and of~\citet{cappart2023combinatorial} for an extended literature review. 
From these surveys, one can observe that most attempts to use learning inside branch-and-bound operate on the branching decisions (i.e., directing the search) or on the primal side (i.e., acting as a heuristic). However, learning to improve the quality of relaxations using better dual bounds has been much less considered. 
This has only been addressed for the restricted use case of solvers based on decision diagrams~\cite{cappart2022improving}, for combinatorial optimization over factor graphs~\cite{deng2022deep},
 integer linear programs~\cite{abbas2024doge}, and
the specific \textsc{weightedCircuit} constraint~\citep{parjadis2023learning}. To our knowledge, learning has never been used for improving CP solvers through Lagrangian decomposition.

\subsubsection{Statement of Contributions.} 

Building on this context, this paper contributes a novel learning approach dedicated to generating tight and valid dual bounds in CP. We propose to achieve this by learning appropriate multipliers for a Lagrangian decomposition, significantly reducing the number of sub-gradient iterations required. 
The learning process is self-supervised, utilizing a graph neural network to represent the problem structure. Unlike most related work, which focuses on learning primal heuristics, our approach develops a learning-based strategy to derive valid and tight dual bounds. To our knowledge, this is the first approach to offer a  generic method for learning valid dual bounds for any type of combinatorial problem.
Experiments are conducted on two challenging case studies: the multi-dimensional knapsack problem and the shift-scheduling problem. 

\section{Lagrangian Decomposition in CP}

Formally, a \textit{constrained optimization problem} (COP) 
is defined as a tuple $\langle X, D(X), C, f \rangle$,
where $X$ is a set of discrete variables, $D(X)$ is the set of domains for
these variables, $C$ is the set of constraints that restrict assignments of values to the variables, and $f(\cdot): X \to \mathbb{R}$ is an objective function.
A feasible solution is an assignment of values from $D(X)$ to $X$ 
that satisfies all the constraints in $C$.
An optimal solution is feasible and also maximizes the objective function. The canonical expression of a COP involving $m$ constraints is formalized below.
\begin{equation}
\label{eq:initial_problem}
\max_{X \in D(X)} \Bigg\{ f(X) ~ \Bigg\vert ~ \bigwedge_{i = 1}^{m} C_i(X) \Bigg \}
\end{equation}

The notation $C_i(X)$ indicates that constraint $C_i$ has the variables $X$ in its scope.
Lagrangian decomposition (LD) involves relaxing this problem to get a valid, ideally tight,  upper bound on the solution. To do so, variables involved in each constraint are duplicated, except for the first constraint. 
\begin{equation}
\label{eq:duplicate_variables}
\max  \Bigg\{ f(X_1) ~ \Bigg\vert ~ \Big ( \bigwedge_{i = 1}^{m} C_i(X_i) \Big) \land  \Big( \bigwedge_{i = 2}^{m} X_i = X_1 \Big)  \Bigg \}
\end{equation}

Here, $X_1$ refers to the initial variables, and  $X_2$ to $X_m$ refers to newly introduced variables.
Next, a penalty term is introduced for each duplicated variable and is 
added to the objective function with a cost $\mu_i \in \mathbb{R}^{|X_i|} \geq 0$.
\begin{multline}
\label{eq:duplicate_variables2}
\max \bigg\{  f(X_1) + \sum_{i=2}^m \mu_i \cdot (X_1 - X_i)    \bigg\vert ~ \\  \Big ( \bigwedge_{i = 1}^{m} C_i(X_i) \Big) \land  \Big( \bigwedge_{i = 2}^{m} X_i = X_1 \Big)  \bigg \} 
\end{multline}

These costs are the \textit{Lagrangian multipliers}. 
We note that the resulting problem retains the same optimal solutions as the initial problem in 
Equation~\eqref{eq:initial_problem}. We now \textit{relax}
the problem by removing all constraints linking the variables.
\begin{equation}
\label{eq:relaxation}
\mathcal{B}(\mu) = \max  \bigg\{ f(X_1) + \sum_{i=2}^m \mu_i \cdot (X_1 - X_i)  ~ \bigg\vert ~  \bigwedge_{i = 1}^{m} C_i(X_i) \bigg \}
\end{equation}

The resulting COP, formalized below, is a relaxation of 
Equation~\eqref{eq:initial_problem} and solving it will provide a dual bound $\mathcal{B}(\mu)$. This corresponds to an upper bound as we are maximizing.
Interestingly, each constraint $C_i$ has a different set of variables and could then be solved independently. We can consequently reorganize the expression as follows.
\begin{align}
\label{eq:final}
\mathcal{B}(\mu) &=  \Phi(X_1)  +  \sum_{i=2}^m \Psi_i (X_i)   \\
\Phi(X_1) &=   \max \bigg \{  f(X_1) +  \Big( \sum_{i=2}^m \mu_i  \Big) \cdot X_1  ~ \Big\vert ~  C_1(X_1)  \bigg \}  \notag \\ 
\Psi_i(X_i) &= \max \big \{ - \mu_i \cdot X_i ~ \big \vert ~  C_i(X_i) \big \} ~  ~ \forall i \in \{2,\dots,m\} \notag 
\end{align}

A valid dual bound can be obtained from a given set of multipliers by solving $m$ independent sub-problems, which are generally easier to solve than the original problem as they involve only a single constraint. However, the quality of the bound (i.e., how close it is to the optimal solution) heavily depends on the values of the multipliers. Finding the optimal multipliers is the main challenge in LD as Equation~\eqref{eq:final} is not differentiable. The most common way is to proceed with a subgradient optimization~\cite {shor2012minimization}. Starting from an arbitrary value, multipliers are iteratively updated.
\begin{equation}
\label{eq:subgradient}
\mu_i^{t+1} = \mu_i^t + \alpha (X^t_1 - X^t_i) ~ ~ ~ ~ \forall i \in \{2,\dots,m\}
\end{equation}

\begin{figure}[ht!]
  \centering

\tikzstyle{object} = [ellipse, minimum width=2.5cm, minimum height=1cm, text centered, draw=black]
\tikzstyle{start} = [ellipse, minimum width=1.5cm, minimum height=1cm, text centered, draw=black]

\tikzstyle{process} = [rectangle, rounded corners, minimum width=2cm, minimum height=1cm, text centered, draw=black, text width=2.8cm]
\tikzstyle{decision} = [diamond, minimum width=1cm, minimum height=1cm, text centered, draw=black]
\tikzstyle{arrow} = [thick,->,>=stealth]
\tikzstyle{dashedarrow} = [thick,->,>=stealth, dashed]
\resizebox{0.47\textwidth}{!} 
{ 
\begin{tikzpicture}
\node (start) [start] {\Large COP};
\node (LD) [process, below of=start, yshift=-1.5cm] {\Large Lagrangian  Decomposition};
\node (decision1) [decision, right of=LD, xshift=1.8cm] {};
\node (SPm) [object, right of=decision1, xshift=1.3cm, yshift=-1.5cm] {\Large$\langle \mu_m, \Phi_m \rangle$};
\node (SP2) [object, right of=decision1, xshift=1.3cm] {\Large$\langle \mu_i, \Phi_i \rangle$};
\node (SP1) [object, right of=decision1, xshift=1.3cm, yshift=1.5cm] {\Large$\langle \mu_1, \Psi \rangle$};

\node (Solve1) [process, right of=SP1, xshift=2.3cm] {\Large$\mathsf{solve}(\Psi) \to X_1$ };
\node (Solve2) [process, right of=SP2, xshift=2.3cm]  {\Large$\mathsf{solve}(\Phi_i) \to X_i$ };
\node (SolveM) [process, right of=SPm, xshift=2.3cm]  {\Large$\mathsf{solve}(\Phi_m): X_m$ };

\node (end) [object, right of=Solve2, xshift=2.2cm] {\Large$\mathcal{B}(\mu)$};

\node (GNN) [start, below of=decision1, yshift=-1.5cm, blue] {\Large$\Omega_\theta(G)$};

\node (TMP) [process, below of=LD , yshift=-1.5cm, blue] {\Large Graph \\ Representation };

\draw [arrow] (start) -- (LD);
\draw [arrow] (LD) -- (decision1);
\draw [arrow] (decision1) -- (SP1);
\draw [arrow] (decision1) -- (SPm);
\draw [arrow] (decision1) -- (SP2);
\draw [arrow] (SP1) -- (Solve1);
\draw [arrow] (SPm) -- (SolveM);
\draw [arrow] (SP2) -- (Solve2);
\draw [arrow] (Solve1) -- (end);
\draw [arrow] (SolveM) -- (end);
\draw [arrow] (Solve2) -- (end);

\draw [dashedarrow, red] (end) to [bend right=75]  node[midway, above] 
{\LARGE$\forall i \in \{2,\dots,m\}: \mu_i^{t+1} = \mu_i^t + \alpha (X^t_1 - X^t_i)$} (decision1);

\draw [dashedarrow, blue] (end) to [bend right=-55]  node[midway, above, xshift=-1.3cm,  yshift=0.7cm] 
{\LARGE$\nabla_\theta \mathcal{B}(\mu) ~~ \text{with} ~~ \mu = \Omega_\theta(G)$} (GNN);

\draw [arrow, blue] (GNN) to node[midway, left] {\Large$\langle \mu_1,\dots,\mu_m \rangle$}  (decision1);
\draw  [arrow, blue] (LD) -- (TMP);
\draw [arrow, blue] (TMP) -- (GNN);
\end{tikzpicture}
}
\caption{Main steps of LD in CP. \textcolor{red}{Red elements} illustrate the standard sub-gradient optimization, while the \textcolor{blue}{blue elements} highlight our contribution based on self-supervised learning.}
\label{fig:LD-learning}
\end{figure}
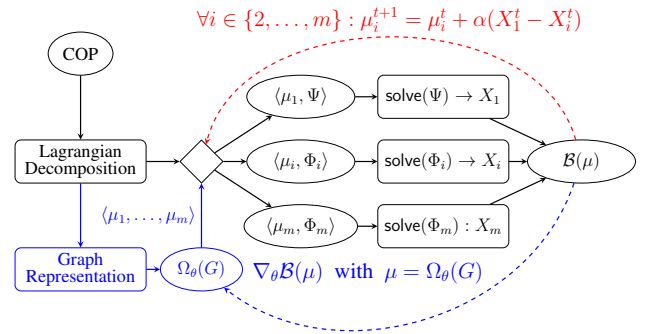

Here, $\mu_i^t$ refers to the set of multipliers $i$ at iteration $t$, $X^t_i$ refers to the optimal values of variables $X_i$ at iteration $t$, and $\alpha$ is a learning rate. The process is illustrated in Figure~\ref{fig:LD-learning} (black and red elements).

\citet{ha2015general} proposed integrating this procedure into the search phase of a CP solver, involving LD and sub-gradient optimization at each node of the tree search. However, this approach is highly time-consuming, as each sub-gradient iteration requires solving each sub-problem, which can be NP-hard. We propose to significantly accelerate this process through the use of learning.

\section{Learning Lagrangian Multipliers}

The bound computed in Equation~\eqref{eq:final} has two notable properties: (1) it can be parameterized using the Lagrangian multipliers $\mu$, and (2) it is always valid, meaning it will never underestimate the actual profit. Both properties present an opportunity to employ a learning-based approach to compute the bound. Inspired by \citet{parjadis2023learning}, we propose building a model $\Omega_\theta : G(V,E) \to \mathbb{R}^{|V|}$ capable of directly predicting all the $\mu$ multipliers for a COP instance given as input and represented by a graph $G$. The graph comprises a set of $V$ nodes (one per multiplier) with their relationships expressed by edges in $E$.
The model $\Omega_\theta$ is parameterized by a set of $p$ parameters $\theta = \{\theta_1, \dots ,\theta_p\}$ and must be differentiable. The goal is to eliminate sub-gradient iterations by directly learning the multipliers that produce a tight bound, thereby saving execution time and enhancing filtering.  The requirement for the model to be differentiable is essential for our methodology, which involves computing the bound using gradient-based optimization.

The goal is to find model parameters $\theta$ 
that yield the lowest possible bound. 
Thanks to the second property, 
the obtained bound is provably valid, regardless of the model's accuracy. 
To our knowledge, this paper presents the first generic method for learning valid dual bounds for any type of discrete COP. This is a notable strength, considering the difficulties in getting guarantees with  learning in combinatorial optimization~\cite{kotary2021end}. Given the multipliers $\mu = \langle \mu_2, \dots, \mu_m \rangle$, the bound minimization problem for an instance $G$ and its related gradient are as follows.
\begin{equation}
\min_\theta \mathcal{B}(\mu) \longmapsto \nabla_\theta \mathcal{B}(\mu) ~~ \text{with} ~~ \mu = \Omega_\theta(G)
\end{equation}

Figure~\ref{fig:LD-learning} illustrates how our procedure (in blue) replaces the standard sub-gradient optimization (in red).
However, computing the gradient of this expression is not trivial, as the bound is obtained through the solutions of combinatorial sub-problems. The chain rule can be applied since the multipliers are parameterized by $\theta$.
\begin{equation}
\nabla_\theta \mathcal{B}(\mu) = \frac{\partial \mathcal{B}(\mu)}{\partial \mu } \cdot \frac{\partial \mu}{\partial \theta}
\end{equation}

The right term, corresponding to the differentiation of the model  $\Omega_\theta$
(i.e., a graph neural network), is easily obtained via backpropagation.
However, the left term requires differentiating the 
expression in Equation~\eqref{eq:final} for all $\mu$.
\begin{equation}
\frac{\partial \mathcal{B}(\mu)}{\partial \mu } =  \frac{\partial \Phi(X_1)}{\partial \mu}  +  \sum_{i=2}^m \frac{\partial \Psi_i (X_i)}{\partial\mu}
\end{equation}

The analytic partial derivative cannot be obtained as it requires solving combinatorial sub-problems. 
Instead, we propose using a locally valid derivative by considering the optimal solution of each sub-problem, allowing us to set all variables $X_i$ for all $i \in \{1, \dots, m\}$. Considering a specific multiplier vector $\mu_i$, we expand the functions $\Phi(.)$ and $\Psi(.)$, which involve only additions of $\mu_i$ and constant multiplications. The partial derivatives of $\mu_i$ are then as follows.
\begin{equation}
\frac{\partial \mathcal{B}(\mu)}{\partial \mu_i } =  (X_1 - X_i) ~ ~ ~ \forall i \in \{2,\dots, m\}
\end{equation}

This yields the gradient in
Equation~\eqref{eq:final}, used to minimize a dual bound obtained through multipliers.
\begin{equation}
\label{eq:gradient_final}
\nabla_\theta \mathcal{B}(\mu) = \bigg \langle  (X_1 - X_i) \cdot \frac{\partial \mu_i}{\partial \theta} ,\dots \bigg \rangle   ~ ~  \forall i \in \{2,\dots, m\}
\end{equation}

We propose leveraging a self-supervised learning approach to parameterize the model $\Omega_\theta$.
The training procedure is formalized in Algorithm~\ref{algo:training-phase}.
It takes as input
a dataset $\mathcal{D}$ consisting of a set of graphs $G(V, E)$ serving as instances of a specific COP.
These  instances can be obtained from historical data or synthetic generation.
The features of the graph are problem-dependent.
At each iteration, LD is carried out, and all the sub-problems are solved through
a dedicated procedure to obtain the optimal values of $X_i$ for current multipliers $\mu$. 
The bound and its gradient are then computed, and a gradient descent step is executed. 
This will change the values of the multipliers for the next
iteration and, consequently, the optimal values of variables $X_i$. 
Finally the parameters $\theta$ of the trained neural network $\Omega_\theta$ are returned.

\begin{algorithm}[!ht]
\begin{algorithmic}[1]

\STATE $\triangleright$ \textbf{Pre:} $\mathcal{D}$ is the set of instances used for training.
            
\STATE $\triangleright$ \textbf{Pre:} $\Omega_\theta$ is the differentiable model to train. 

\STATE $\triangleright$ \textbf{Pre:} $\theta$ are randomly initialized parameters.

\STATE $\triangleright$ \textbf{Pre:} $K$ is the number of training epochs.

~

\FOR{$k \mathbf{~from~} 1 \mathbf{~to~} K$}

\STATE $\langle G,m \rangle := \mathsf{sampleFromTrainingSet}(\mathcal{D})$

\STATE $\langle \Phi,\Psi_2,\dots,\Psi_m \rangle := \mathsf{LD}(G)$

\STATE $X_1 := \mathsf{solve}(\Phi)$

\FOR{$i \mathbf{~from~} 2 \mathbf{~to~} m$}

\STATE $X_i := \mathsf{solve}(\Psi_i)$

\ENDFOR

\STATE  $\mu := \Omega_\theta(G)$

\STATE $\mathcal{B}(\mu) := \Phi(X_1) + \sum_{i=2}^m \Psi(X_i)$

\STATE $\theta := \theta - \nabla_\theta \mathcal{B}(\mu)  $

\ENDFOR

\RETURN{$\theta$}

\end{algorithmic}
\caption{Training phase of $\Omega_\theta$ from a dataset $\mathcal{D}$.}
\label{algo:training-phase}
\end{algorithm}

\section{Graph Neural Network Architecture}

\textit{Graph neural networks} (GNNs)~\citep{scarselli2008graph} are a specific neural 
architecture designed to compute a vector representation, 
also referred to as an \textit{embedding}, for each node of an input graph. 
This architecture has been widely used in related works on learning for combinatorial optimization~\cite{cappart2023combinatorial}.
Learning is carried out by aggregating information from neighboring nodes multiple times. 
Each aggregation operation is referred to as a \textit{layer} of the GNN 
and involves weights that must be learned.

Let $G(V, E)$ be a simple graph used as input of $\Omega_\theta$ in Algorithm~\ref{algo:training-phase}, 
let $h_i \in \mathbb{R}^p$ be a $p$-dimensional vector representation of the input features of a node $v \in V$, and let $k_{u,v} \in \mathbb{R}^q$ be a similar $q$-dimensional representation of 
the input features of an edge $(u,v) \in E$. Finally, let $L$ be the number of layers of the GNN.
Our implementation is based on \textsc{ResGatedConv} architecture~\citep{bresson2017residual}.
The inference process involves computing the next representation 
($h^{l+1}_v$) from the previous one ($h^{l}_v$) 
for each node $v$. This process is formalized in Equation~\eqref{eq:gnn1},
where $\langle \theta_1^l, \dots, \theta_4^l \rangle$ are the weights in layer 
$l$, $\mathcal{N}(v)$ are the neighboring nodes of $v$, 
$\odot$ is the element-wise product, 
$\eta_{v,u}$ is an edge gate between nodes $v$ and $u$, 
and $\sigma$ is a sigmoid activation. 
\begin{align}
\label{eq:gnn1}
        h_v^{l+1} &= \textsc{ReLU} \Big( \theta_1^l h_v^l + \sum_{u \in \mathcal{N}(v)} \left( \eta_{v,u} \odot \theta_2^l \right) h_u^l \Big) \\
\eta_{v,u} &= \sigma \Big( \theta_3^l h_v^l + \theta_4^l h_u^l \Big) 
\end{align}

The edge features $k_{u,v}$ are concatenated to each message.
Applying the inference on the $L$ layers results in an embedding $h_v^{L}$ for each node $v \in V$. Similarly, we also compute a \textit{graph embedding} $\Gamma \in \mathbb{R}^r$ by averaging information from each node, as shown in Equation~\eqref{eq:gnn2}. 
\begin{equation}
\label{eq:gnn2}
\Gamma = \frac{1}{|V|} \sum_{v \in V} \sigma ( h_v^{L} )
\end{equation}

Considering both a graph and node embedding allows each node to combine a global and a local representation of the problem. Finally, both embeddings are concatenated ($\cdot \| \cdot$) and given as input of a fully-connected neural network (\textsc{FCNN}) to obtain the Lagrangian multipliers, as shown in Equation~\eqref{eq:gnn4}. We recall that there is one node per multiplier in our representation.
The architecture has two messages passing steps with a  size of 64 each and the fully-connected neural network has two layers with 256 and 128 neurons.
\begin{equation}
\label{eq:gnn4}
\mu_v = \textsc{FCNN} \big( h_v^{L} ~ \big \| ~  \Gamma  \big) ~ ~ \forall_v \in \{1,\dots,V\}
\end{equation}

\section{Case studies}

\begin{table*}[!ht]
\resizebox{\textwidth}{!} 
{ 
\centering
\renewcommand{\arraystretch}{1.3}
\begin{tabular}{l rrr rrr rrr}\toprule
\multicolumn{10}{c}{Multi-dimensional Knapsack Problem (MKP)} \\
\midrule
\multirow{2}{*}{Approaches} & \multicolumn{3}{c}{30 items} & \multicolumn{3}{c}{50 items}  & \multicolumn{3}{c}{100 items} \\
\cmidrule(lr){2-4} \cmidrule(lr){5-7} \cmidrule(lr){8-10}
 & No. solved  & Time (sec.) & No. nodes & No. solved  & Time (sec.) & No. nodes & No. solved  & Time (sec.) & No. nodes \\
\midrule
\midrule
\textsf{CP} & 50/50 & \textbf{0.5} & 22,000 & 30/50 &1,100 & 40,000,000  & 0/50 & - & - \\
\textsf{CP+SG} & 50/50 & 19.2 & 116 & 50/50 & 158 & 436 & 41/50 & 2.4K & 4.1K \\
\midrule
 \textsf{CP+Learning($all$)} & 50/50 & 2.0 & 235 & 50/50 & \textbf{36} & 2,600 & 25/50 & 2.8K & 146K \\
\textsf{CP+Learning($all$)+SG} & 50/50  & 6.7 & 170 & 50/50  & 40 & 1,700 & 29/50 & 1.6K & 33K \\
\textsf{CP+Learning($root$)+SG}& 50/50  & 10.6 & \textbf{81} & 50/50  & 83 & \textbf{340} & \textbf{49/50} & \textbf{1.1K} & \textbf{2.1K} \\
\midrule
\midrule
\multicolumn{10}{c}{Shift Scheduling Problem (SSP)} \\
\midrule
\multirow{2}{*}{Approaches} & \multicolumn{3}{c}{10 symbols and 20 states} & \multicolumn{3}{c}{20 symbols and 20 states}  & \multicolumn{3}{c}{10 symbols and 80 states} \\
\cmidrule(lr){2-4} \cmidrule(lr){5-7} \cmidrule(lr){8-10}
 & No. solved  & Time (sec.) & No. nodes & No. solved  & Time (sec.) & No. nodes & No. solved  & Time (sec.) & No. nodes \\
\midrule
\midrule
\textsf{CP+SG} & \textbf{28/50} & 473 & 2.0K & \textbf{27/50} & 750 & \textbf{2.9K} & 13/50 & 3.3K & \textbf{2.5K}  \\
\midrule
 \textsf{CP+Learning($all$)} & 24/50 & 852  & 8.8K & 24/50  & \textbf{660} & 4.7K & \textbf{20/50} & \textbf{1.3K} & 3.3K  \\
\textsf{CP+Learning($all$)+SG} & 24/50 & 880 & 6.3K & 22/50 & 770 & 4.5K & 17/50 & 1.6K & 3.1K  \\
\textsf{CP+Learning($root$)+SG}& \textbf{28/50} & \textbf{453} & \textbf{1.8K} & 22/50 & 1,100 & 4.5K & 13/50 & 3.4K & \textbf{2.5K}  \\
\bottomrule
\end{tabular}
}
\caption{Comparison of our approaches with standard baselines for the MKP and SSP on 50 instances per configuration. Results include the number of instances solved within a 2-hour timeout, average execution time, and average nodes explored to close the search. Averages are calculated only on instances solved by each method (excluding \textsf{CP} for MKP with 100 items). The \textsf{CP} baseline is also omitted for SSP due to its significantly poorer performance.}

\label{tab:main-results}

\end{table*}

\begin{figure*}[ht!]
  \centering

\begin{subfigure}[b]{0.30\textwidth}
\includegraphics[width=\textwidth]{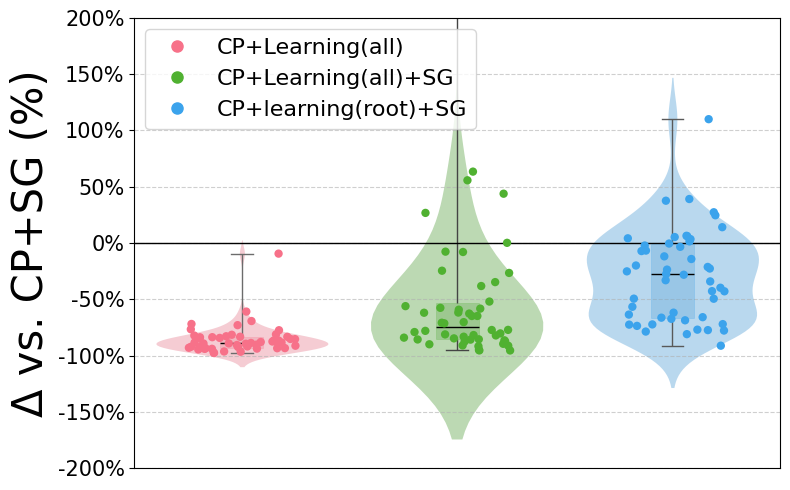}
\caption{MKP with 30 items.}
\label{subfig:box-1}
\end{subfigure}
\hfill
\begin{subfigure}[b]{0.30\textwidth}
\includegraphics[width=\textwidth]{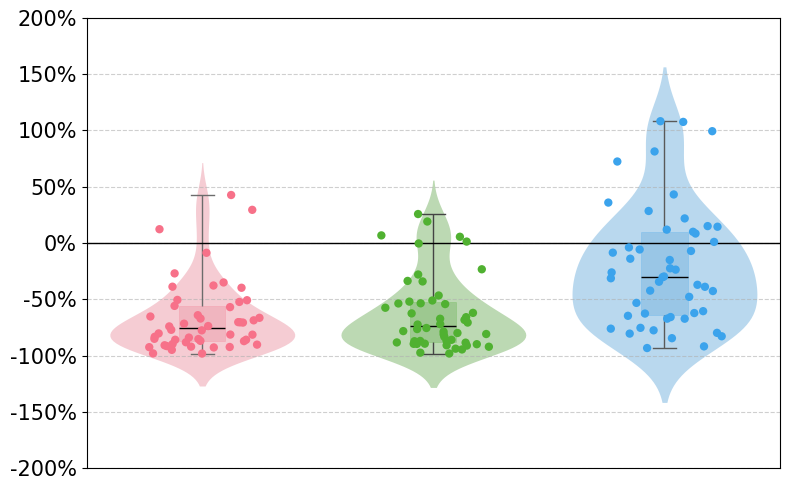}
\caption{MKP with 50 items.}
\label{subfig:box-2}
\end{subfigure}
\hfill
\begin{subfigure}[b]{0.30\textwidth}
\includegraphics[width=\textwidth]{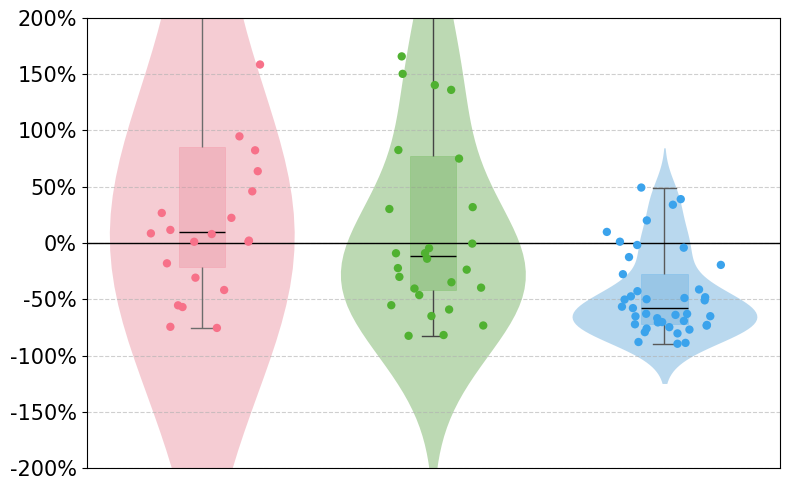}
\caption{MKP with 100 items.}
\label{subfig:box-3} 
\end{subfigure}
\begin{subfigure}[b]{0.30\textwidth}
\includegraphics[width=\textwidth]{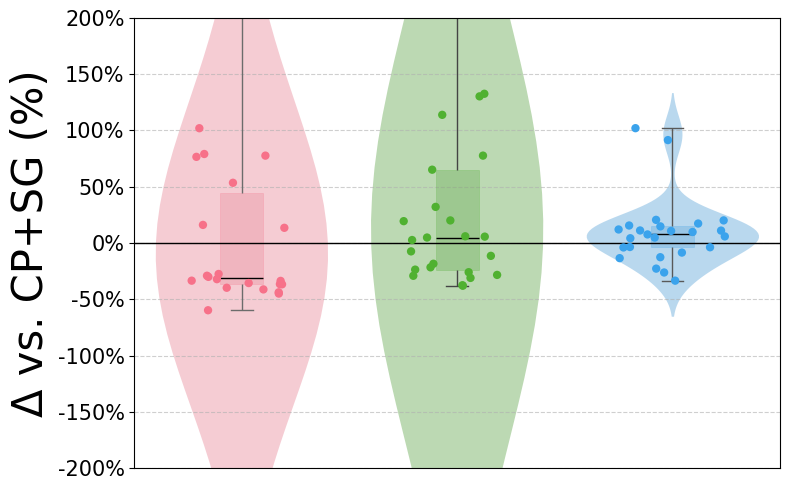}
\caption{SSP with 10 symbols and 20 states.}
\label{subfig:box-4}
\end{subfigure}
\hfill
\begin{subfigure}[b]{0.30\textwidth}
\includegraphics[width=\textwidth]{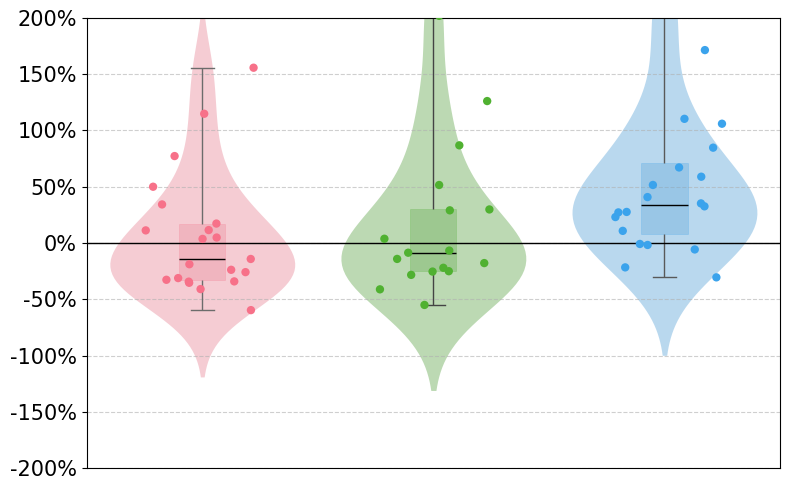}
\caption{SSP with 20 symbols and 20 states.}
\label{subfig:box-5}
\end{subfigure}
\hfill
\begin{subfigure}[b]{0.30\textwidth}
\includegraphics[width=\textwidth]{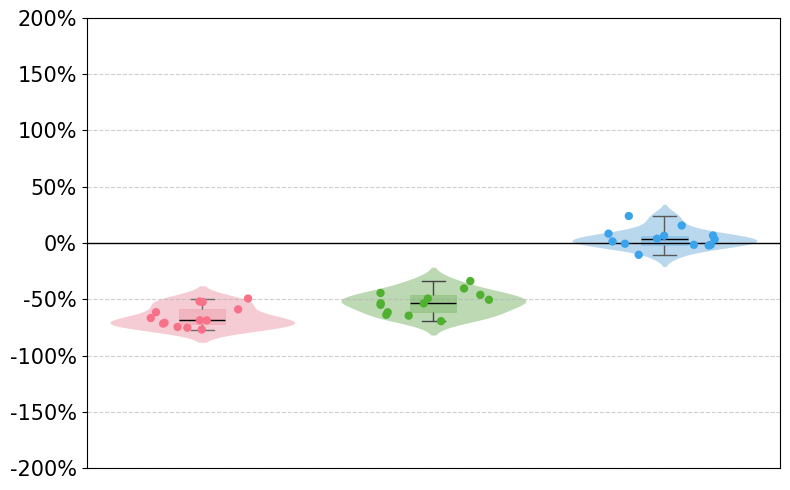}
\caption{SSP with 10 symbols and 80 states.}
\label{subfig:box-6}
\end{subfigure}

  \caption{Relative performance of our approaches compared to the \textsf{CP+SG} baseline in execution time. Each dot below 0\% indicates a reduction in execution time with our method for a specific instance.} 
  \label{fig:box-plot}
\end{figure*}

This paper addresses two challenging COPs: the multi-dimensional knapsack and the shift scheduling problem, chosen for direct comparison with \citet{ha2015general}. Further details are in the appendix.

\subsubsection{Multi-dimensional Knapsack Problem (MKP).} It is an extension of the classic knapsack problem. In the MKP, we are given a set of items, each with a specific profit and multiple weights (one per dimension). Additionally, there are multiple knapsack constraints, each representing a capacity limit for a different dimension. The objective is to select a subset of items that maximizes the total profit while ensuring that the total weight in each dimension does not exceed the respective capacity constraints. In our CP formalization, for an instance of dimension $d$, we have $d$ specific knapsack constraints, which result in $d$ sub-problems. We solve each sub-problem with a standard dynamic programming algorithm for one-dimensional knapsack, which runs in $\mathcal{O}(nW)$, with $n$ the number of items and $W$ the capacity.
    For training and evaluation, a synthetic dataset is generated following the protocol of Pisinger~\cite{Kellerer2004}. Instances of $30$, $50$, and $100$ items and up to 5 dimensions are considered. Profits follow a uniform distribution between 0 and 500, 
    weights between 0 and 100, and the capacity is a percentage of the total weight per dimension.

\subsubsection{Shift Scheduling Problem (SSP).} The problem involves assigning employees to shifts in a way that meets work regulations while maximizing profit. We consider a challenging version where the work regulations are so complex that modeling them for a single employee requires numerous \textsc{Regular} constraints~\citep{pesant2004regular}. The goal is to assign one activity per shift such that the transitions between activities over the periods satisfy the constraints and maximize the total profit. We consider instances with two \textsc{Regular} constraints (yielding two sub-problems) and 50 periods, as they are already challenging to solve and cannot all be solved to optimality within a 2-hour timeout.
The sub-problems are solved using the procedure introduced by \citet{pesant2004regular}, which runs in $\mathcal{O}(nmQ)$, where $n$ is the number of variables, $m$ is the number of symbols, and $Q$ is the number of states in the \textsc{Regular} automaton. 
A synthetic dataset is generated for training and evaluation, following the procedure of \citet{pesant2004regular}. Instances with two constraints, 50 items, and three configurations of symbols and states ($10/20$, $10/80$, $20/20$) are considered, where $x/y$ denotes a configuration with $x$ symbols and $y$ states. Profits follow a uniform distribution between 0 and 100, with 0.3 as the proportion of undefined transitions and 0.5 as the proportion of final states.

\section{Experimental Protocol}

This section outlines the  protocol used to evaluate the efficacy and reliability of our approach, detailing the datasets and  configurations used across various testing scenarios.

\subsubsection{Competitors.} The methods tested in the experiments are:
\begin{itemize}
\item \textsf{CP}. A pure constraint programming approach without Lagrangian decomposition or learning.
\item \textsf{CP+SG}. The same CP model is enhanced with LD and sub-gradient optimization as proposed by \citet{ha2015general}.
We re-implemented this approach, and our results for this baseline are much better than those reported in the initial paper, likely due to the improved branching heuristics and the choice of CP solver.

\item \textsf{CP+Learning($all$)}. The CP model with LD, using the bounds learned in Alg.~\ref{algo:training-phase} instead of sub-gradient, with learning applied at every node of the search tree.
\item \textsf{CP+Learning($all$)+SG}. Similar to the previous approach, the learned bounds are further improved by sub-gradient optimization. The trained model $\Omega_\theta$ is called at every search tree node to obtain the bounds.
\item \textsf{CP+Learning($root$)+SG}. Model $\Omega_\theta$ is used only at the root node, with the resulting bound serving as the initial value for bootstrapping sub-gradient  in the other nodes.
\end{itemize}

\subsubsection{Implementation.}

The graph neural network has been implemented in 
\textit{Python} using \textit{Pytorch Geometric}~\cite{fey2019fast}. 
The CP models are  solved with \textit{Gecode}~\citep{schulte2006gecode},
a  CP solver implemented in \textit{C++}.
The interface between both languages has been implemented using  \textit{libtorch}. The implementation and the datasets used are released on GitHub\footnote{Hidden while the work is under review. Available on demand.}.

\subsubsection{Training and Validation.}

The training phase follows
Algorithm~\ref{algo:training-phase}. 
For each configuration, a specific model is trained using synthetic instances. A training set of 900 instances is used for the MKP and 200 for the SSP, with data augmentation from partially solved instances up to a depth of 5. The self-supervised learning process eliminates the need for labeled multipliers. Training was conducted on a Tesla V100-SXM2-32GB GPU for 24 hours, with no extensive hyperparameter tuning (details in the attached files). Models were trained and tested in a single run, with convergence observed on a validation set of 100 MKP and 30 SSP instances. Adam optimizer~\citep{kingma2014adam} with 
a learning rate of $0.001$ has been used.

\section{Empirical Results}

This section presents the results of our experimental analysis, offering a detailed quantitative evaluation of our approach through a series of experiments.

\subsubsection{Main Result: Quality of the Learned Bounds.}
Table~\ref{tab:main-results} shows the performance of all approaches on a test set of 50 instances for both problems. We first observe that CP without LD, although being the fastest in the easiest configuration (MKP with 30 items), has a very large number of explored nodes, which quickly explodes for larger instances.
Interestingly,
we observe that learning alone (\textsf{CP+Learning($all$})) consistently results in a larger search tree (i.e., more nodes) compared to sub-gradient optimization (\textsf{CP+SG}), indicating weaker bounds. However, this approach allows the solver to skip all sub-gradient iterations, often leading to significant execution time savings (e.g., from 158 to 36 seconds for MKP with 50 items, or from 3,300 to 1,300 seconds for SSP with 80 states). Unfortunately, these time savings are insufficient for the most challenging configurations, where the number of nodes is much higher.
In such cases, using learning to initialize the sub-gradient procedure (\textsf{CP+Learning($root$)+SG}) offers the best performance (e.g., reducing nodes explored from 4,100 to 2,100 for MKP with 100 items, decreasing average execution time from 2,400 to 1,100 seconds, and solving 8 additional instances). In conclusion, learning has significantly improved the application of LD in CP, either by replacing sub-gradient steps or by bootstrapping them. Performance on individual instances 
(Figure~\ref{fig:box-plot}) supports this conclusion. The significance of each improvement in time was validated using a one-sided Wilcoxon signed-rank test. For each  configuration, at least one of our approaches offers a significant improvement ($p$-value $< 0.001$), excepting SSP-20-20 ($p$-value $< 0.1$) and SSP-10-20 (no significant improvement).

\subsubsection{Analysis: Fine-Tuning to OOD Instances.}
This experiment evaluates our approach's ability to generalize to instances from an unseen distribution when fine-tuning is applied. We used the standard benchmark of \citet{shih1979branch}, consisting of 30 MKP instances with 30 to 90 items. Fine-tuning was limited to 2 hours (10 epochs) with 100 instances from a similar distribution. Results in Table~\ref{tab:OOD-MKP} show that we outperform the baseline for most instances, thanks to fine-tuning. Although \textsf{CP+Learning($all$)} offers the best overall performance, it falls short in a few cases due to the intrinsic lack of guarantees in learning-based methods. Combining learning with sub-gradient (\textsf{CP+Learning($root$)+SG}) mitigates this risk and provides more stable results. We also report that performance without fine-tuning  is worse than the baseline.

\subsubsection{Analysis: Evolution of the Bounds.}

\begin{figure}[ht!]
  \centering

\begin{subfigure}[b]{0.50\columnwidth}
\includegraphics[width=\textwidth]{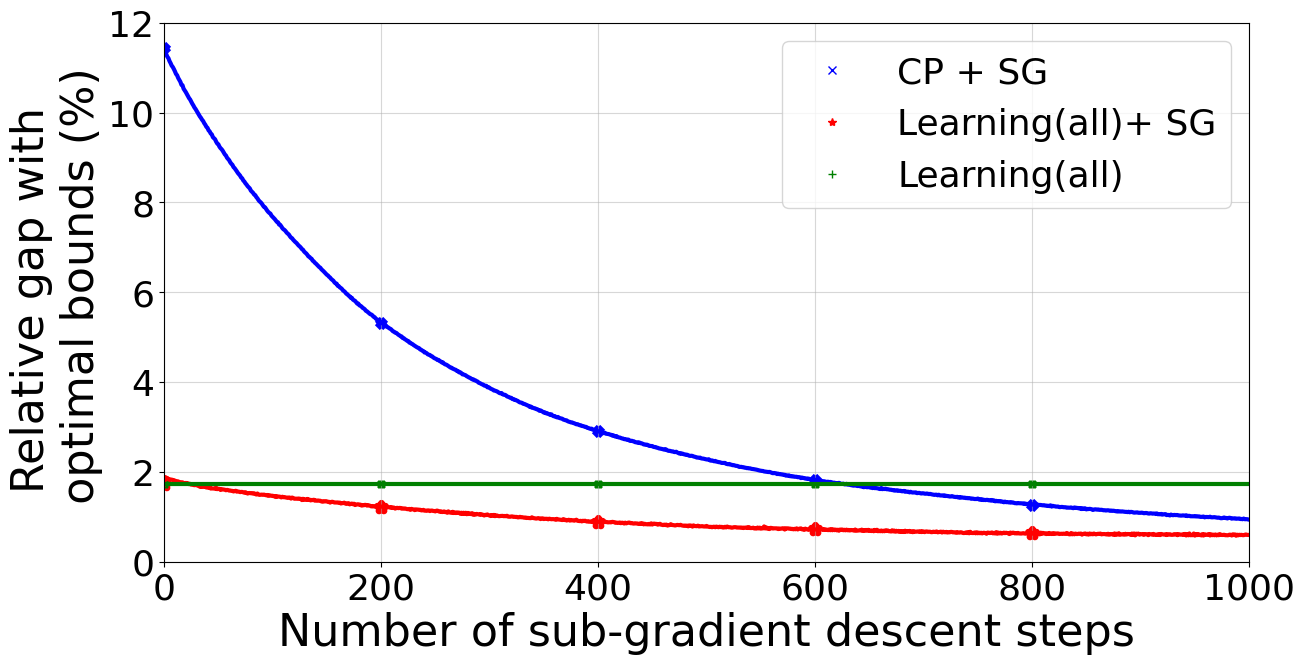}
\caption{MKP (100 items).}
\label{subfig:bound-1}
\end{subfigure}
\hfill
\begin{subfigure}[b]{0.47\columnwidth}
\includegraphics[width=\textwidth]{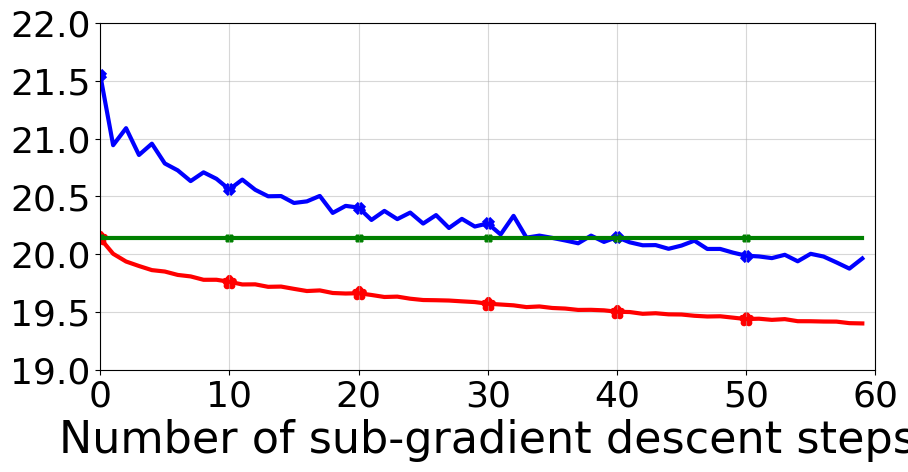}
\caption{SSP (10 vars., 80 states).}
\label{subfig:bound-2}
\end{subfigure}
  \caption{Evolution of the bounds for the two most challenging configurations (gap with the optimal solution).} 
  \label{fig:bound-progress}
\end{figure}

This experiment analyzes the tightness of bounds obtained 
with different approaches and how they improve with each sub-gradient step. Figure~\ref{fig:bound-progress} shows the average bounds on 50 instances for the two hardest configurations, measured by the gap with the optimal solution. In both cases, learning alone provides much better multipliers than random ones (LD before any sub-gradient step). After a number of iterations (600 for MKP and 40 for SSP), sub-gradient alone  manages to  improve the bounds but is still outperformed by our approach, which uses learning to initialize the sub-gradient procedure.

\section{Conclusion}

Previous works have identified Lagrangian decomposition as a promising approach for developing automatic bounding mechanisms in constraint programming. However, the practicality of LD is limited by the prohibitive execution time required for sub-gradient optimization at each node of the tree search. This paper addresses this challenge with an innovative approach that learns Lagrangian multipliers to produce tight dual bounds. The self-supervised learning process, which eliminates the need for labeled bounds, leverages a graph neural network to represent the problem structure. Experiments demonstrate the potential of this approach, either by replacing sub-gradient iterations entirely or by bootstrapping them. Fine-tuning has shown promise for generalizing to out-of-distribution instances. To our knowledge, this is the first contribution proposing a generic method for learning valid dual bounds in constraint programming.

\begin{table}[!ht]
\resizebox{\columnwidth}{!} 
{ 
\centering
\renewcommand{\arraystretch}{1.3}
\begin{tabular}{ll rr rr rr}\toprule

\multicolumn{2}{c}{Instances} & \multicolumn{2}{c}{\textsf{CP+SG}} & \multicolumn{2}{c}{\textsf{CP+Learning($all$)}} & \multicolumn{2}{c}{\textsf{CP+Learning($root$)+SG}} \\

\cmidrule(lr){1-2} \cmidrule(lr){3-4} \cmidrule(lr){5-6}  \cmidrule(lr){7-8} 

Name &  Vars & Nodes & Time & Nodes & Time & Nodes & Time (seconds)  \\
\midrule
\textsf{Weish1}	&	30		&	29	&	6	&	341	&	3	&	29	&	4	\\
\textsf{Weish2}	&	30		&	41	&	7	&	159	&	1	&	41	&	6	\\
\textsf{Weish3}	&	30		&	51	&	6	&	67	&	$<$ 0	&	43	&	5	\\
\textsf{Weish4}	&	30		&	29	&	4	&	35	&	$<$ 0	&	29	&	3	\\
\textsf{Weish5}	&	30		&	25	&	4	&	31	&	$<$ 0	&	25	&	3	\\
\textsf{Weish6}	&	30		&	51	&	14	&	649	&	6	&	51	&	12	\\
\textsf{Weish7}	&	30		&	43	&	13	&	1197	&	10	&	43	&	11	\\
\textsf{Weish8}	&	30		&	51	&	15	&	669	&	6	&	51	&	13	\\
\textsf{Weish9}	&	30		&	43	&	11	&	61	&	1	&	41	&	9	\\
\textsf{Weish10}	&	50		&	117	&	72	&	291	&	4	&	101	&	36	\\
\textsf{Weish11}	&	50		&	69	&	30	&	73	&	1	&	45	&	13	\\
\textsf{Weish12}	&	50		&	85	&	38	&	323	&	4	&	65	&	23	\\
\textsf{Weish13}	&	50		&	139	&	56	&	379	&	5	&	107	&	33	\\
\textsf{Weish14}	&	60		&	153	&	84	&	623	&	10	&	127	&	53	\\
\textsf{Weish15}	&	60		&	111	&	42	&	581	&	6	&	111	&	37	\\
\textsf{Weish16}	&	60		&	97	&	40	&	393	&	5	&	87	&	31	\\
\textsf{Weish17}	&	60		&	337	&	223	&	9127	&	126	&	337	&	177	\\
\textsf{Weish18}	&	70		&	427	&	210	&	70564	&	1307	&	429	&	186	\\
\textsf{Weish20}	&	70		&	177	&	94	&	1159	&	15	&	171	&	75	\\
\textsf{Weish21}	&	70		&	237	&	115	&	991	&	16	&	233	&	97	\\
\textsf{Weish22}	&	80		&	167	&	174	&	1207	&	33	&	147	&	119	\\
\textsf{Weish23}	&	80		&	373	&	376	&	2729	&	77	&	299	&	231	\\
\textsf{Weish24}	&	80		&	501	&	309	&	t.o.	&	t.o.	&	511	&	281	\\
\textsf{Weish25}	&	80		&	307	&	186	&	7444	&	140	&	319	&	170	\\
\textsf{Weish26}	&	90		&	233	&	259	&	t.o.	&	t.o.	&	197	&	158	\\
\textsf{Weish27}	&	90		&	269	&	216	&	t.o.	&	t.o.	&	261	&	175	\\
\textsf{Weish28}	&	90		&	297	&	311	&	1465	&	49	&	253	&	193	\\
\textsf{Weish29}	&	90		&	347	&	353	&	1599	&	56	&	303	&	205	\\
\textsf{Weish30}	&	90		&	1015	&	1062	&	t.o.	&	t.o.	&	1085	&	997	\\
\midrule
\multicolumn{2}{l}{Arithmetic mean} & 203 & 153 & -& -& 191  & 117  \\
\multicolumn{2}{l}{Geometric mean} & 131 & 60 &- & -& 120  & 44  \\
\multicolumn{2}{l}{No. best} & 5/30 & 0/30 & 0/30 & \textbf{25/30} & \textbf{17/30}  & 5/30  \\
\bottomrule

\end{tabular}
}
\caption{Fine-tuning on MKP instances \citep{shih1979branch}. \textit{No. best} is how often each approach achieved the best results.}

\label{tab:OOD-MKP}

\end{table}

\bibliography{aaai25}
\newpage
\appendix

\section{Multi-Dimensional Knapsack Problem}
The \textit{Multi-Dimensional Knapsack Problem} (MKP)  is an extension of the classic knapsack problem~\citep{Kellerer}. In the MDKP, there are multiple constraints (or dimensions) rather than just one. Each item $i$ has a value $v_i$ and multiple weights $\{w^1_i, \dots,w^D_i\}$, each corresponding to a different dimension $d$. The goal is to select a subset of items that maximizes the total value while ensuring that the sum of the weights in each dimension does not exceed the capacity $W^d$ of that dimension. This problem has numerous practical applications, including resource allocation, budgeting, and portfolio optimization, where multiple constraints must be simultaneously satisfied. 

\subsection{Constraint Programming Model}

Let $x_i$ be a binary variable indicating if the item $i$ is inserted into the knapsack.
This problem can be easily formalized as a constrained optimization problem.
\begin{flalign}
\max_x ~& \sum_{i=1}^{n} v_i x_i & \\
\text{s. t.} ~ &  \sum_{i=1}^{n} w^d_{i} x_i \leq W^d & \forall d\in \{1,\dots,D\} \notag \\
& x_i \in \{0,1\} & \forall i \in \{1,\dots,n\} \notag
\end{flalign}

\subsection{Graph Encoding}

We propose the following encoding to represent a MKP instance into a graph $G(V,E)$.
We recall that in our representation, 
each node $v \in V$ corresponds to a specific Lagrangian multiplier. 
A node $v$ is created for each pair of variable and constraint present in the problem instance. Then, an instance of $n$ items and $d$ dimensions yields a graph with $n\times d$ nodes. Nodes are linked by edges if they share the same variable or constraint. Each node $v$ is also decorated with six features:

\begin{itemize}
    \item $h^1_v$: the index of the related variable.
    \item $h^2_v$: the index of the related constraint. 
    \item $h^3_v$: the profit of the item. 
    \item $h^4_v$: the weight of the item on the related dimension.
    \item $h^5_v$: the ratio of profit to weight.
    \item $h^6_v$: the ratio of weight to capacity.
\end{itemize}





\section{Shift Scheduling Problem}
The \textit{Shift-Scheduling Problem} (SSP) involves scheduling employees' activities (e.g., work, break, lunch, rest) while adhering to labor regulations. Constraint programming is well-suited for solving this problem, as complex work regulations can be naturally formalized using \textsc{Regular} constraints~\cite{pesant2004regular}. These constraints are typically specified using a non-deterministic finite automaton $\mathcal{A}$, defined by the 5-tuple $\mathcal{A} = (Q, A, \delta, q_0, F)$ where:
\begin{itemize}
    \item \( Q \) is a finite set of states;
    \item \( A \) is the alphabet;
    \item \( \delta : Q \times A \rightarrow Q \) is the transition function;
    \item \( q_0 \in Q \) is the initial state;
    \item \( F \subseteq Q \) is the set of final states.
\end{itemize}

The constraint $\textsc{Regular}(\langle x_1,\dots,x_n \rangle,\mathcal{A})$ is satisfied if the sequence of variable values $\langle x_1, \ldots, x_n \rangle$ belongs to the regular language recognized by a deterministic finite automaton $\mathcal{A}$. Intuitively, each variable $x_i$ represents an employee's activity at time $i$. The constraint is satisfied if if the sequence of states complies with the transition function and leads to a final state in the automaton. We consider in this paper a simple yet challenging version of the SSP with one employee, complex labor regulations, and multiple regular constraints.

\subsection{Constraint Programming Model}

Let $W$ be the set of work activities, $T$ the set of periods, $\mathcal{A}$ the set of deterministic finite automata representing labor regulations, and $p_{ij}$ (from a table $P$) the profit from assigning activity $i \in W$ to period $j \in T$. Each period must have an assigned activity  that respects the automata transitions and maximizes total profit.
Let $x_i$ be a decision variable representing the activity assigned to period $j$, and let $\textsc{Element}(i,T,v)$ be a constraint that holds if $v = T[i]$. The SSP model is formulated as proposed by \citet{ha2015general}. Specifically, $\textsc{Element}(x_j, P, c_j)$ ensures that the profit $c_j$ equals $p_{i,j}$ if and only if $x_j = i$.
\begin{flalign}
\max_x ~& \sum_{j \in T} c_j & \\
\text{s. t.} ~ &  \textsc{regular}(\langle x_1,\dots,x_t\rangle, \mathcal{A}_i) & \forall \mathcal{A}_i \in \mathcal{A} \notag \\
& \textsc{element}(x_j, p_{i,j}, c_j) & \forall i \in W, \forall j \in T \notag \\
& x_j \in W & \forall j \in T \notag \\
& \min_{i \in W} p_{ij} \leq c_j \leq \max_{i \in W} p_{ij} & \forall j \in T \notag \\
& c_j \in \mathbb{R} &  \forall j \in T   \notag  
\end{flalign}

\subsection{Graph Encoding}

We propose the following encoding to represent a SSP in-
stance into a graph $G(V, E)$.  A node $v$ is created for each triplet of (variable, constraint, value) present in the problem instance. Then,
an instance of $n$ variables, $m$ constraints, and $d$ values inside the domain yields
a graph with $n\times m \times d$ nodes. 
Nodes are linked by edges if they correspond to the same (variable, value) pair or the same constraint, subject to a valid transition defined by the automata. Each node is decorated with four features:
\begin{itemize}
    \item $h^1_v$: the index of the related variable.
    \item $h^2_v$: the index of the related constraint. 
    \item $h^3_v$: the index of the related value. 
    \item $h^4_v$: the profit associated with the triplet.
\end{itemize}

\end{document}